\documentclass[letterpaper]{article} 
\usepackage[preprint]{aaai2027}  
\usepackage[hyphens]{url}  
\usepackage{graphicx} 
\usepackage{natbib}  
\usepackage{caption} 
\usepackage{algorithm}
\usepackage{algorithmic}

\usepackage{newfloat}
\usepackage{listings}
\DeclareCaptionStyle{ruled}{labelfont=normalfont,labelsep=colon,strut=off} 
\floatstyle{ruled}
\newfloat{listing}{tb}{lst}{}
\floatname{listing}{Listing}

\usepackage{booktabs}

\usepackage{url}
\usepackage{multirow}
\usepackage{graphicx}
\usepackage{subcaption}
\usepackage[accsupp]{axessibility}  
\usepackage[table]{xcolor}

\newcommand{\methodname}{HyperGS}

\title{HyperGS: Fast and Generalizable Gaussian Video Representation}

\author{
    Fatimah Zohra\textsuperscript{\rm 1}, Chen Zhao\textsuperscript{\rm 1,}\textsuperscript{\rm 2}, Shuming Liu\textsuperscript{\rm 1}, Yahya Al Malallah\textsuperscript{\rm 1}, Bernard Ghanem\textsuperscript{\rm 1}
}
\affiliations{
    \textsuperscript{\rm 1} King Abdullah University of Science and Technology (KAUST)\\
    \textsuperscript{\rm 2} Harbin Institute of Technology, Shenzhen \\
}

\begin{document}

\maketitle

\begin{abstract}
Gaussian Splatting has emerged as an effective representation for video, but existing methods rely on per-video optimization. This leads to slow encoding and limits generalization across videos. To amortize this optimization, we propose \methodname{}, a feedforward, optimization-free approach that directly predicts Gaussian representations from any video in a single forward pass, speeding up encoding and decoding by orders of magnitude while generalizing to out-of-distribution videos at higher resolutions. In \methodname{}, we design a factorized spatiotemporal Transformer to extract tokens  from video, and a learnable query-based Transformer  to obtain 8-parameter Gaussian representations for each video frame. We find that naively predicting Gaussians across diverse videos induces a needle-like degeneration that collapses training, and address this with a rank-based geometric regularizer whose strength adapts dynamically to stabilize optimization. \methodname{} achieves encoding at $10^4$--$10^5\times$ the speed of per-video Gaussian optimization at matched reconstruction quality while generalizing zero-shot to $720p$ video, enabling higher-resolution rendering without re-encoding. \methodname{} improves PSNR by +2.9--3.1\,dB over the prior video encoders on K400, SSv2, and UCF101 at a smaller video representation size. By predicting explicit 2D Gaussians in a single forward pass, \methodname{} combines the fast, flexible rendering of Gaussian Splatting with the speed and generalization of feedforward prediction, advancing Gaussians as a practical direction for fast and generalizable video representation.
\end{abstract}

\section{Introduction}

Recent advances in 3D reconstruction have increasingly been introduced into the video representation field, changing how video data is encoded and reconstructed. Among these, 3D Gaussian Splatting (3DGS)~\cite{kerbl20233d} has emerged as a particularly effective representation, modeling a scene as a collection of explicit, differentiable primitives that can be rendered directly through rasterization rather than through an implicit, coordinate-based function. The adaptation of 3DGS to dynamic video in particular has brought improvements in rendering speed and geometric stability. Yet state-of-the-art 3DGS video models~\cite{bond2025gaussianvideo, liu2025d2gv, sun2024splatter} remain bound by a fundamental limitation: \textit{per-video optimization}. Because each video's Gaussians are fit independently through iterative, gradient-based optimization---often tens of thousands of steps and on the order of hours per video---none of the structure learned while encoding one video transfers to the next. This makes per-video Gaussian methods impractical for real-time processing.

A feedforward, optimization-free alternative instead amortizes this encoding by training a shared model to predict a video's representation in a single forward pass. Prior work along this line predicts the \textit{weights} of an implicit neural decoder, tying the representation to a specific neural architecture. We show that predicting explicit primitives instead decouples the representation from any implicit decoder, extending the efficient rendering and flexibility of Gaussian Splatting to a feedforward regime.

Our choice of Gaussians as the predicted representation follows directly from this decoupling. Gaussian parameters reduce the regression target into geometrically meaningful subproblems---position, scale, orientation, color---providing a natural inductive bias for the prediction task. Rendering via a sum of Gaussians is also a direct learning objective, since the image is composed linearly from the predicted primitives, while an implicit decoder maps predicted weights to the reconstruction through an additional nonlinear step.

To show this, we propose \methodname{}, a feedforward, optimization-free approach that predicts explicit Gaussian Splatting representations directly from video in a single forward pass, following the hypernetwork paradigm of predicting a representation rather than optimizing it. \methodname{} pairs a factorized spatiotemporal encoder with a query-based decoder that aggregates video features into per-frame Gaussian parameters via learnable weight tokens, producing dynamic, independent primitives for each frame that capture the spatial and temporal dynamics of a video. The predicted Gaussians are rendered through differentiablerasterization~\cite{zhang2024gaussianimage}, and the entire model is trained end-to-end with a reconstruction loss. Our contributions are as follows:
\begin{itemize}
    \item We propose \methodname{}, the first feedforward, optimization-free method to predict explicit  Gaussian primitives for video in a single forward pass. \methodname{} improves PSNR by +2.9--3.1\,dB over prior video encoders on K400, SSv2, and UCF101, using a smaller video representation size while encoding at $\sim$400--2200 FPS and decoding at $\sim$3900--4400 FPS for 3000--250 Gaussians per frame respectively.
    \item \methodname{} generalizes zero-shot to unseen, high-resolution video, encoding $10^4$--$10^5\times$ faster than per-video Gaussian optimization at matched reconstruction quality on 720p UVG and DAVIS. A patch-grid encoding scales this further, tiling a high-resolution frame into a grid of patches---trading a larger Gaussian count and longer encoding for higher fidelity.
    \item We identify a training instability in feedforward-predicted Gaussians where unconstrained optimization drives Gaussians toward needle-like anisotropy, causing mid-training collapse and introduce an adaptive effective-rank regularizer with EMA-tracked percentile thresholds to stabilize training.
\end{itemize}

\section{Related Works}

\subsection{Gaussian Splatting for Video Representation}
3D Gaussian Splatting~\cite{kerbl20233d} introduced an efficient alternative to NeRF for 3D scene modeling, replacing volumetric ray marching with differentiable rasterization of explicit Gaussian primitives. It has since been adapted for static 2D image representation~\cite{zhang2024gaussianimage,zhang2025image} and extended to video by adding temporal dynamics. Video approaches differ mainly in how they parameterize motion, falling into two broad categories. The first builds on 2D Gaussian Splatting and uses temporal networks to deform flat splats across frames, as in D2GV~\cite{liu2025d2gv} and STGV~\cite{lin2026stgv}. The second treats the scene as inherently 3D, modeling motion holistically through explicit trajectories or learned deformation fields. Trajectory-based methods define explicit motion paths for each Gaussian over time. GaussianVideo~\cite{bond2025gaussianvideo} attaches cubic B-spline trajectories to 3D Gaussians and uses Neural ODEs for continuous camera dynamics. VGR~\cite{sun2024splatter} combines polynomial and Fourier bases for motion, regularized by distilling 2D priors like optical flow and monocular depth into an editable canonical space. Deformation-based methods instead keep a single canonical set of Gaussians and learn a neural network to deform them at each timestep. 4D-GS~\cite{wu20244d} uses 4D neural voxels to encode spatial-temporal features decoded by a lightweight MLP. D-3DGS~\cite{yang2024deformable} relies on a purely implicit MLP with an annealing smoothing strategy to counter noisy camera tracking.

\subsection{Hypernetworks for Implicit Neural Representation}
Implicit Neural Representations represent a video as a neural network mapping coordinates to pixel values~\cite{sitzmann2020implicit,chen2021nerv,chen2023hnerv,li2022nerv}, achieving competitive performance with fast decoding, but at the cost of expensive per-video encoding. A line of work instead predicts the decoder's weights directly in a single forward pass via a hypernetwork. TransINR~\cite{chen2022transformers} jointly processes input patch tokens and learnable weight tokens via Transformer self-attention to produce the full parameter set, while GINR~\cite{kim2023generalizableimplicitneuralrepresentations} improves parameter efficiency by restricting prediction to a single early decoder layer and sharing the remainder across instances. FastNeRV~\cite{chen2024fast} predicts weights through a ViT-based network with layer-adaptive weight token allocation and a parallelized group-convolution decoder to achieve fast encoding and decoding speeds. TeCoNeRV~\cite{padmanabhan2026teconerv} addresses the quadratic memory scaling that limited prior methods to $256\times 256$ by decomposing weight prediction into resolution-independent patch tubelets. It further reduces bitstream size by storing only weight residuals between consecutive clips and applying temporal-coherence regularization to shrink those residuals, enabling results at 480p--1080p for the first time in this line of work.

\subsection{Conventional Video Compression}
Standard video compression relies on hybrid codecs that reduce spatial and temporal redundancy through a multi-stage pipeline of block-based prediction, transform coding, quantization, and entropy coding. Standards from H.264/AVC~\cite{wiegand2003overview} to H.265/HEVC~\cite{sullivan2012overview} refine this pipeline's prediction and partitioning schemes to improve rate-distortion performance. While these codecs achieve strong compression ratios, their block-based, sequentially-predicted bitstreams are tied to a fixed resolution, non-differentiable, and require decoding preceding frames in a GOP before accessing a target frame or region — precluding continuous rendering, gradient-based editing, and true random access.

\section{Method}
\label{sec:method}
\begin{figure}[t]
\centering
\footnotesize
\includegraphics[width=0.5\textwidth]{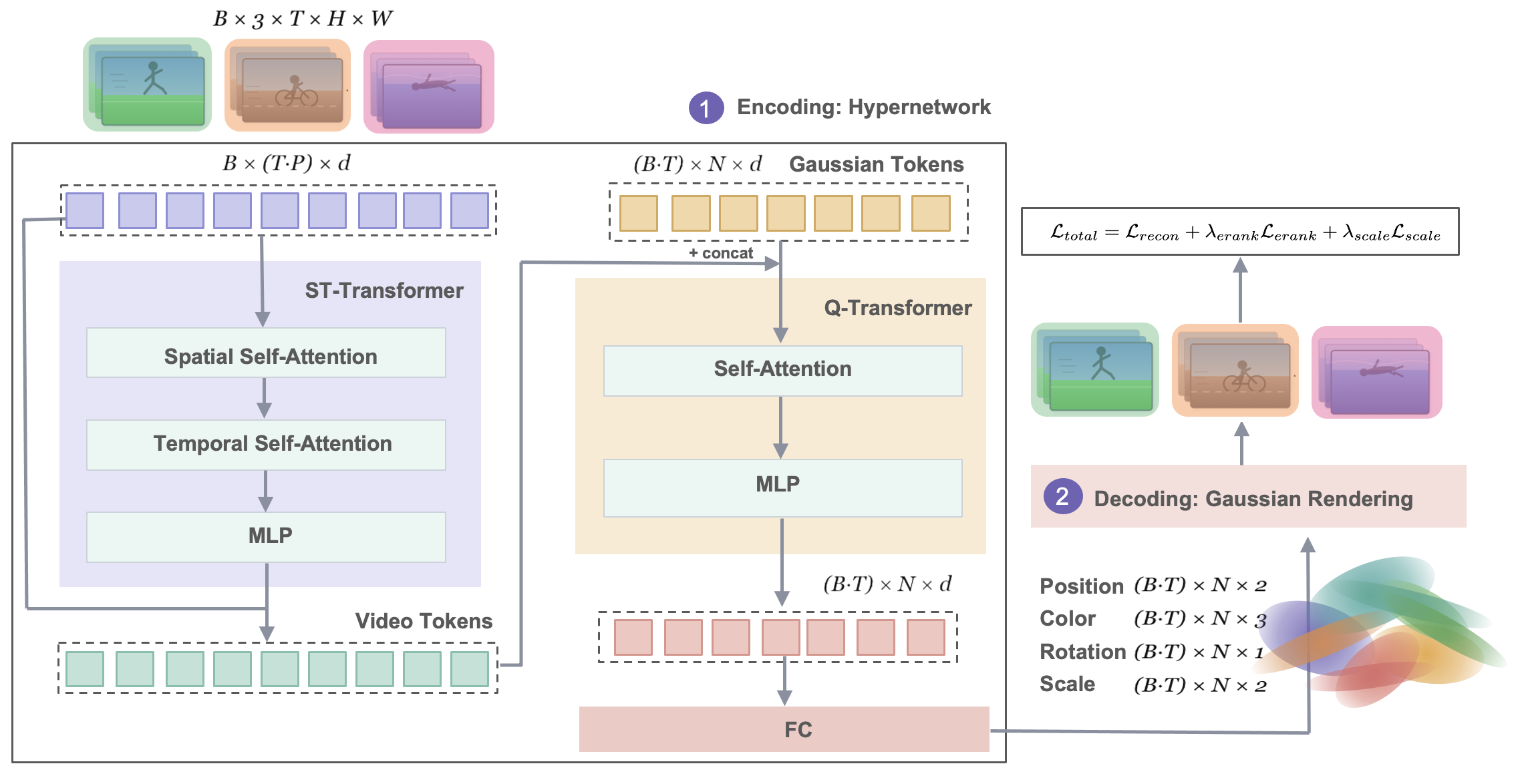}
\caption{\methodname{} maps input videos directly to Gaussian primitives with a single transformer hypernetwork. A factorized \textbf{ST-Transformer} processes video tokens with spatial and temporal self-attention. Learnable Gaussian tokens then attend to these features via the learnable queries of the \textbf{Q-Transformer} and are directly regressed into per-frame Gaussian parameters (position, scale, rotation, color), which are rasterized to reconstruct each frame.}

\label{fig:arch}
\end{figure}

We learn a network $f_\theta$ that maps an input video $V \in \mathbb{R}^{B \times 3 \times T \times H \times W}$ directly to a temporal sequence of explicit 2D Gaussian representations $\mathcal{G} = \{\mathcal{G}_1, \ldots, \mathcal{G}_T\}$ in a single forward pass. Each set $\mathcal{G}_t$ corresponds to the representation of frame $t$ and consists of $N$ geometric primitives. Specifically, the $i$-th Gaussian in frame $t$ is defined by a tuple of interpretable parameters: mean position $\boldsymbol{\mu}_i \in \mathbb{R}^2$, anisotropic scale $\mathbf{s}_i \in \mathbb{R}^2$, rotation $r_i \in [0, 2\pi)$, and color $\mathbf{c}_i \in \mathbb{R}^3$.

\subsection{Video Tokenization}
We tokenize the video $V$ using a 3D convolutional projection with a kernel size of $1 \times p \times p$ (tubelet size of 1) and a learnable spatio-temporal positional embedding. This maps the pixel space to a sequence of patch embeddings $\mathbf{Z}^0 \in \mathbb{R}^{B \times (T \cdot P) \times d}$, where $P = (H \cdot W)/p^2$ denotes the number of spatial patches per frame. Furthermore, relative position information is dynamically injected via Rotary Position Embeddings (RoPE) during the subsequent encoder attention layers.

\subsection{Spatio-Temporal Hypernetwork} 
\methodname{} processes video tokens and a set of learnable Gaussian tokens within a single network. The video tokens are first contextualized across space and time by a Factorized Spatio-Temporal Transformer (ST-Transformer). A set of learnable Gaussian tokens is then introduced and processed by a Query Transformer (Q-Transformer), where they attend to the contextualized video features and are directly regressed into Gaussian parameters.

\paragraph{ST-Transformer.}
Each ST-Transformer layer $\ell$ applies spatial self-attention with 2D RoPE to encode grid coordinates $(h, w)$, temporal self-attention with 1D RoPE based on frame indices $t$, and a feedforward network sequentially:
\begin{align}
\mathbf{Z}^{\ell}_{\text{spat}} &= \mathbf{Z}^{\ell-1} + \operatorname{SpatialAttn}_{\text{2D}}(\operatorname{LN}(\mathbf{Z}^{\ell-1})) \\
\mathbf{Z}^{\ell}_{\text{temp}} &= \mathbf{Z}^{\ell}_{\text{spat}} + \operatorname{TemporalAttn}_{\text{1D}}(\operatorname{LN}(\mathbf{Z}^{\ell}_{\text{spat}})) \\
\mathbf{Z}^{\ell} &= \mathbf{Z}^{\ell}_{\text{temp}} + \operatorname{FFN}(\operatorname{LN}(\mathbf{Z}^{\ell}_{\text{temp}}))
\end{align}

This factorization reduces the computational complexity of the Transformer to $\mathcal{O}(T \cdot P^2 + P \cdot T^2)$. 

After $L_{\text{enc}}$ layers, we apply a skip connection summing the input embeddings with the final ST-Transformer's output:
\begin{equation}
\mathbf{Z}_{\text{out}} = \mathbf{Z}^{L_{\text{ST}}} + \mathbf{Z}^0
\end{equation}
The skip connection preserves the original patch embeddings, allowing the Q-Transformer to access both temporally-contextualized features and frame-specific local features. We find that this skip connection helps stabilize the training, albeit trading-off stability for performance (see supplement for details).

\paragraph{Q-Transformer.}
The Q-Transformer translates the dense feature map produced by the ST-Transformer into a sparse set of Gaussian parameters via learnable queries. The ST-Transformer output is reshaped to the per-frame view and concatenated with $N$ learnable \textbf{Gaussian tokens} $\mathbf{W} \in \mathbb{R}^{N \times d}$ where $N$ denotes the number of Gaussian points representing each frame:
\begin{equation}
\mathbf{X}^0_{\text{Q}} = [\mathbf{Z}_{\text{frame}};\ \mathbf{W}]
\end{equation}

The Q-Transformer consists of $L_{\text{Q}}$ self-attention layers. Through self-attention, the weight tokens aggregate information from the temporally contextualized frame-level patches to localize features relevant for reconstruction:
\begin{align}
\mathbf{X}^{\ell}_{\text{Q}} &= \mathbf{X}^{\ell-1}_{\text{Q}} + \operatorname{SelfAttn}(\operatorname{LN}(\mathbf{X}^{\ell-1}_{\text{Q}})) \\
\mathbf{X}^{\ell}_{\text{Q}} &= \mathbf{X}^{\ell}_{\text{Q}} + \operatorname{FFN}(\operatorname{LN}(\mathbf{X}^{\ell}_{\text{Q}}))
\end{align}

\paragraph{Gaussian Parameter Prediction.}
The transformed gaussian tokens $\mathbf{O}_w = \mathbf{X}^{L_{\text{Q}}}_{\text{Q}}[:, -N:, :] \in \mathbb{R}^{(B \cdot T) \times N \times d}$ serve as the latent representations for the Gaussian primitives. These tokens are projected to predict the final Gaussian parameters and are constrained to valid ranges prior to rasterization. We map positions $\boldsymbol{\mu} \in \mathbb{R}^{N \times 2}$ with a $\tanh$ activation to the normalized range $[-1, 1]$ and colors $\mathbf{c} \in \mathbb{R}^{N \times 3}$ with a sigmoid to $[0, 1]$. Rotations $\mathbf{r} \in \mathbb{R}^{N \times 1}$ use a sigmoid scaled by $2\pi$ to cover the full angular range $[0, 2\pi]$. Scales $\mathbf{s} \in \mathbb{R}^{N \times 2}$ are enforced positive via $s_i = |\hat{s}_i| + \epsilon$ to avoid degenerate Gaussians. We use a fixed opacity $\alpha=1.0$.

\subsection{Gaussian Rasterization}
We use the 2DGS \cite{zhang2024gaussianimage} approach, in which the primitives are projected by computing the 2D covariance matrix $\boldsymbol{\Sigma}$ from the predicted scale and rotation. We use the rotation-scale decomposition which guarantees positive semi-definiteness:
\begin{equation}
\boldsymbol{\Sigma} = \mathbf{R}(r) \begin{pmatrix} s_x^2 & 0 \\ 0 & s_y^2 \end{pmatrix} \mathbf{R}(r)^\top
\end{equation}
where $\mathbf{R}(r)$ is the standard 2D rotation matrix constructed from $\cos(r)$ and $\sin(r)$. The final video frames are reconstructed via a differentiable 2D Gaussian rasterizer. Given the fixed opacity, each pixel intensity is simply an additive signal accumulation. The intensity at pixel location $\mathbf{p}$ is given by:
\begin{equation}
I(\mathbf{p}) = \sum_{i=1}^{N} \mathbf{c}_i \cdot \exp\left(-\frac{1}{2}(\mathbf{p} - \boldsymbol{\mu}_i)^\top \boldsymbol{\Sigma}_i^{-1} (\mathbf{p} - \boldsymbol{\mu}_i)\right)
\end{equation}

\subsection{Training Objective}
Our model is trained end-to-end to reconstruct the target video frames while maintaining well-conditioned geometric primitives. The overall training objective $\mathcal{L}_{total}$ is formulated as a combination of a photometric reconstruction loss and two adaptive geometric regularization penalties:
\begin{equation}
\mathcal{L}_{total} = \mathcal{L}_{recon} + \lambda_{erank} \mathcal{L}_{erank} + \lambda_{scale} \mathcal{L}_{scale}
\end{equation}
where $\lambda_{erank}$ and $\lambda_{scale}$ are scalar weights controlling the influence of the regularization terms.

\paragraph{Reconstruction Loss.} To measure the photometric fidelity of the rendered frames against the ground truth video, we employ a standard Mean Squared Error (MSE) loss. For a video sequence of length $T$, the loss is computed over all spatial pixel coordinates $\mathbf{p} = (x, y)$ for each frame $t$:
\begin{equation}
\mathcal{L}_{recon} = \frac{1}{T \cdot H \cdot W} \sum_{t=1}^{T} \sum_{\mathbf{p}} \left\| I_t(\mathbf{p}) - \hat{I}_t(\mathbf{p}) \right\|_2^2
\end{equation}
where $I_t(\mathbf{p})$ is the intensity rendered by the additive Gaussian rasterizer for frame $t$, and $\hat{I}_t(\mathbf{p})$ is the corresponding ground truth pixel intensity.

\paragraph{Adaptive Covariance Regularization.}
Unconstrained optimization of 2D Gaussians leads to ``needle-like'' artifacts—Gaussians that are extremely stretched along one axis (i.e., $s_x \gg s_y$). We regularize shape quality using the Effective Rank regularization introduced in \cite{hyung2024effectiverankanalysisregularization}, which treats the normalized squared scales as a probability distribution $q_k = s_k^2 / \sum_j s_j^2$ and computes $\operatorname{erank} = \exp(H(q))$, where $H(q) = -\sum_k q_k \log q_k$ is the Shannon entropy. For 2D Gaussians, $\operatorname{erank}$ ranges from $1$ (a degenerate needle) to $2$ (a perfect circle).

However, setting the regularization thresholds is sensitive to model capacity, making manual tuning difficult. We address this with adaptive thresholds that track the distribution of predicted Gaussian shapes during training and penalize only the most extreme outliers. We observe that adaptive regularization provides consistent training stability across all settings, but can overconstrain higher-capacity models (e.g., $\geq$3000 Gaussians, $\geq$16 frames), limiting their representational flexibility. In these cases, fixed thresholds can achieve higher PSNR by allowing the optimizer greater freedom to exploit useful elliptical Gaussians.

The adaptive regularization maintains an exponential moving average (EMA) of the 1st-percentile of the effective rank across the batch:
\begin{equation}
\tau_{erank}^{(t)} = \max\!\Big(\tau_{\text{floor}},\; m \cdot \tau_{erank}^{(t-1)} + (1-m) \cdot \operatorname{P}_{0.01}\!\big[\operatorname{erank}(\mathcal{G})\big]\Big)
\end{equation}
where $m$ is the EMA momentum (we use $m=0.99$) and $\tau_{\text{floor}}$ is a hard lower bound (e.g., $\tau_{\text{floor}} = 1.005$) that prevents the threshold from drifting too close to $1$ under collective degeneration. Only Gaussians whose effective rank falls below the adaptive threshold are penalized via a hinge loss:
\begin{equation}
\mathcal{L}_{erank} = \frac{1}{|\mathcal{G}|} \sum_{i \in \mathcal{G}} \max\!\left(0,\; \tau_{erank}^{(t)} - \operatorname{erank}_i\right)
\end{equation}

\paragraph{Adaptive Scale Magnitude Regularization.}
We similarly prevent scales from growing unreasonably large by tracking the 99th-percentile of the predicted scale magnitudes with an EMA:
\begin{equation}
s_{max}^{(t)} = \min\!\Big(s_{\text{ceil}},\; m \cdot s_{max}^{(t-1)} + (1-m) \cdot \operatorname{P}_{0.99}\!\big[|\mathbf{s}(\mathcal{G})|\big]\Big)
\end{equation}
where $s_{\text{ceil}}$ is a hard upper bound (e.g., $s_{\text{ceil}} = 80$) that prevents the threshold from drifting upward. Scale values exceeding the adaptive threshold are penalized with a squared excess loss:
\begin{equation}
\mathcal{L}_{scale} = \frac{1}{|\mathcal{G}|} \sum_{i \in \mathcal{G}} \max\!\left(0,\; s_i - s_{max}^{(t)}\right)^2
\end{equation}

\section{Experiments}
\label{sec:experiments}

We compare against two types of baselines: (i) ~\textit{per-video optimization} GS methods that fit a representation to each video independently and (ii) ~\textit{feedforward methods} that predict representations in a single forward pass.
For the GS methods, we primarily compare \methodname{} with  2DGS~\cite{zhang2024gaussianimage} since \methodname{} can be viewed as the amortized counterpart to 2DGS, as both predict the same structure of 2D Gaussians but differ in how the parameters are obtained---feed-forward prediction versus iterative optimization. We also compare with temporally adapted 2DGS methods D2GV~\cite{liu2025d2gv} and STGV~\cite{lin2026stgv}. Both adapt 2DGS to video by applying frame-wise deformations to a canonical set of 2D Gaussians, though they differ in their underlying mechanisms.

\paragraph{Training Details.}
We train on a subset of Kinetics-400~\cite{zisserman2017kinetics} (25 videos per class) following~\cite{chen2024fast}, using the LAMB optimizer~\cite{you2020largebatchoptimizationdeep} with a learning rate of $1{\times}10^{-3}$, cosine annealing over 400 epochs, and FP16 mixed precision on 4$\times$A100 GPUs. All videos are resized to $256{\times}256$ resolution for training. Unless otherwise noted, all encoding/decoding speeds are measured with a per-GPU batch of $B{=}8$ at $256{\times}256$.

\paragraph{Evaluation Details.}
Kinetics-400 (K400)~\cite{zisserman2017kinetics} serves as the in-distribution test set. To assess zero-shot generalization, we evaluate on four out-of-distribution benchmarks without any fine-tuning: Something-Something V2 (SSv2)~\cite{goyal2017something}, UCF-101~\cite{UCF101}, UVG~\cite{mercat2020uvg} and DAVIS 2016~\cite{davis2016}. We primarily evaluate at $256{\times}256$ by resizing videos on the shorter side and center-cropping. For K400, SSv2, and UCF-101, we uniformly sample 8 frames per video. We evaluate a subset of DAVIS 2016 \footnote{Dance, Camel, Bmx, Swan, Cow, Boat as in \cite{liu2025d2gv} and \cite{lin2026stgv}} and UVG at 720p with 4$\times$ sub-sampling applied to UVG.

\paragraph{Higher-Resolution Rendering.} Because our Gaussian representation is continuous and resolution-independent, it can be rasterized at any target resolution at inference time without re-encoding. The ground truth is prepared at the target resolution (ex. $1280{\times}720$) by resizing on the shorter side and center-cropping. The ST-Transformer always receives a fixed $256{\times}256$ input, and the predicted Gaussians are rendered \emph{directly} at the target resolution. For K400, UCF-101, and SSv2, whose native resolutions fall below the target, both ground truth and prediction are upsampled, so the comparison mainly reflects content fidelity at higher display resolutions.

\label{sec:patch-grid}
\paragraph{Patch-Grid Encoding.} Because the network is trained to map an encoder view to Gaussians independently of any particular video or resolution, it can be applied repeatedly across a scene to reach resolutions beyond its native input size. This matters because a single $256 \times 256$ encoder view of a high-resolution frame requires aggressive downscaling that discards fine detail. For this reason, we optionally tile the target frame into a $g_c \times g_r$ grid of $256 \times 256$ patches, encode each patch independently with the same netowrk in a single batched forward pass, render every tile, and combine them back into the full frame. This trades representation size---a total of $g_c \, g_r \times N$ Gaussians per frame---for quality at constant network capacity, and still requires no per-video optimization.

\label{sec:results}

\begin{figure}[t]
  \centering
  \begin{subfigure}[b]{\linewidth}
    \centering
    \includegraphics[width=0.85\linewidth]{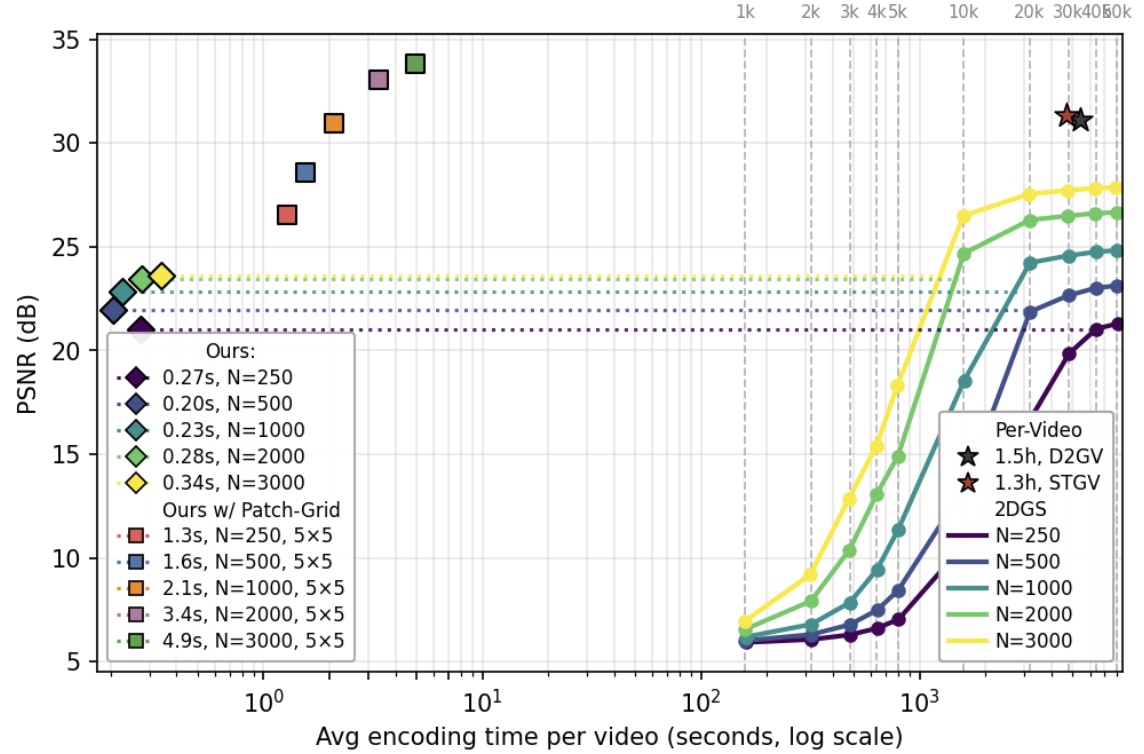}
    \caption{Davis 2016 at 720p, $\sim$82 frames/video.}
    \label{fig:davis-encoding-psnr}
  \end{subfigure}
  \begin{subfigure}[b]{\linewidth}
    \centering
    \includegraphics[width=0.85\linewidth]{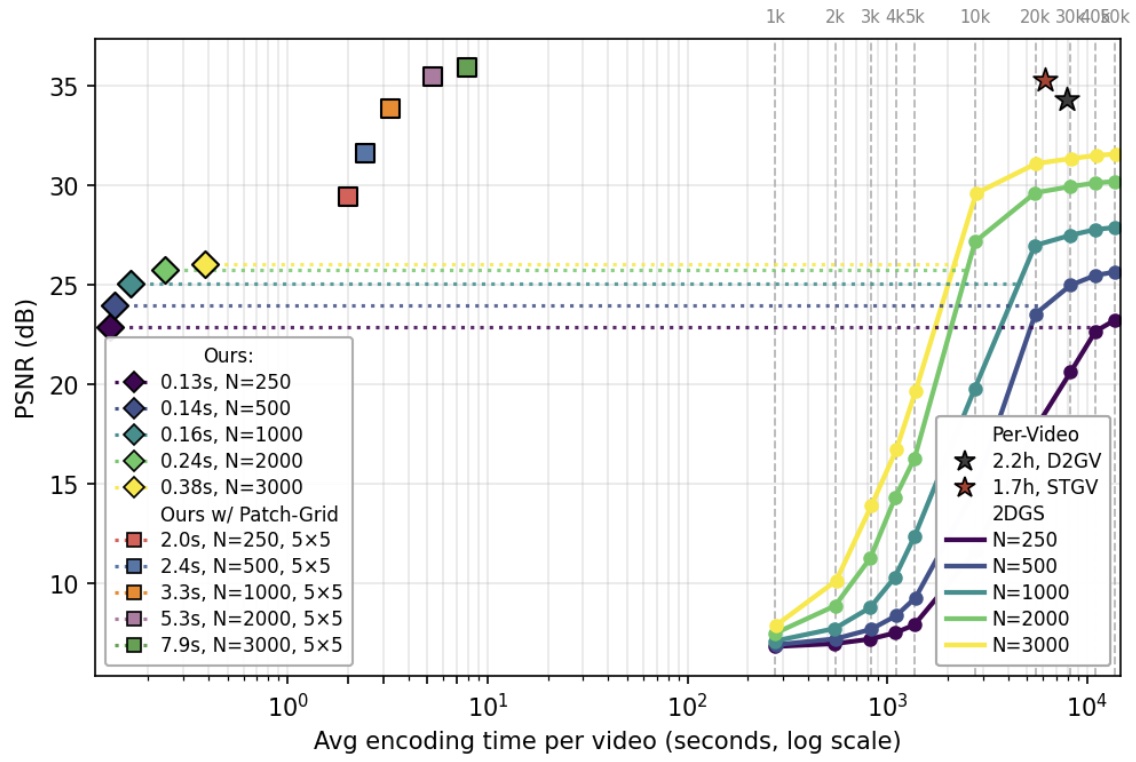}
    \caption{UVG at 720p. $\sim$139 frames/video.}
    \label{fig:second-figure-label}
  \end{subfigure}
  \caption{ Each curve shows 2DGS~\cite{zhang2024gaussianimage} trained \emph{per-frame} with a fixed number of Gaussians (250--3000). The curve plots PSNR against the average \emph{per-video} encoding time at different training iterations ($1k-50k$).
  } 
  \label{fig:davis-uvg-encoding-psnr}
\end{figure}

\paragraph{\textbf{Comparison with Per-Video Optimization.}}
Figure~\ref{fig:davis-uvg-encoding-psnr} compares \methodname{} against its non-amortized counterpart, 2DGS, in terms of encoding speed and reconstruction quality, evaluated zero-shot at 720p on DAVIS 2016 and UVG. Each 2DGS curve corresponds to a fixed Gaussian count $N$ optimized independently per frame for a given number of iterations (up to 50k, requiring on the order of hours per video)---more iterations yield a tighter per-frame fit and higher PSNR, as expected. \methodname{} instead produces its Gaussians in a single forward pass, encoding an entire video in milliseconds. Encoding time for our methods is the GPU pipeline consisting of a single forward pass plus Gaussian rasterization. At matched reconstruction quality, this reduces encoding time by $10^4$--$10^5\times$ relative to 2DGS, despite \methodname{} never having trained on these videos and resolutions.

We additionally evaluate a patch-grid variant (Sec.~\ref{sec:patch-grid}) that closes the remaining quality gap entirely. At inference time, we tile each frame into in-distribution $256\times256$ patches and encode them independently in a single batched forward pass, trading a larger representation size and longer encoding time for improved reconstruction quality. This surpasses per-video optimization on both datasets: on UVG, a $5\times5$ grid at $N{=}3000$ reaches 35.93\,dB in $\sim$7.9s per video while on DAVIS, a $5\times5$ grid at $N{=}3000$ reaches 33.84\,dB in $\sim$4.9s per video. This flexibility is inherent to feedforward prediction, whereas per-video optimization methods can only infer at the exact resolution they were trained on and lack the ability to operate over a patch grid at inference time. Across both datasets, quality improves monotonically with Gaussian count $N$ in both modes, and patch encoding offers the best quality--compute trade-off whenever full-frame PSNR falls short.

\begin{table*}[t]
\centering
\setlength{\tabcolsep}{0.5mm}
\footnotesize
\begin{tabular}{l cc cc cc cc cc}
\toprule
\multirow{2}{*}{Method}
 & Model & Repr. \#Params & Enc. & Dec.
 & \multicolumn{2}{c}{K400} & \multicolumn{2}{c}{SSV2} & \multicolumn{2}{c}{UCF} \\
 \cmidrule(lr){6-7} \cmidrule(lr){8-9} \cmidrule(lr){10-11}
 & \#Params & / Video & time (s)$\downarrow$ & VPS$\uparrow$
 & PSNR$\uparrow$ & SSIM$\uparrow$ & PSNR$\uparrow$ & SSIM$\uparrow$ & PSNR$\uparrow$ & SSIM$\uparrow$ \\
\midrule
\multicolumn{11}{l}{\textit{Feedforward methods at} 256 $\times$ 256 $\times$ 8} \\
\midrule
TransINR & 48.0\,M & 99\,K & 0.01 & ---
  & 20.3 & 0.595 & 22.8 & 0.703 & 20.7 & 0.591 \\
GINR& 47.6\,M & 139.4\,K & 0.01 & ---
  & 22.8 & 0.65 & 25.3 & 0.737 & 22.7 & 0.651 \\
FastNeRV & 47.6\,M & 85.6\,K & 0.01 & $\text{\textbf{4.6}}\times\text{\textbf{10}}^3$
  & 27.4 & 0.78 & 30.3 & 0.835 & 26.8 & 0.778 \\
\rowcolor[gray]{0.9} \textbf{\methodname{}} ($N{=}1000$) & 47.7\,M & \textbf{64\,K} & \textbf{0.007} & $0.51\!\times\!10^3$
  & \textbf{30.3} & \textbf{0.867} & \textbf{33.4} & \textbf{0.898} & \textbf{29.8} & \textbf{0.871} \\
\midrule
\multicolumn{11}{l}{\textit{Feedforward methods at} 640 $\times$ 480 $\times$ 8} \\
\midrule
FastNeRV$^\dagger$ & --- & 436.8K & 0.080 & $0.57\!\times\!10^3$
  & 22.86 & 0.721 & — & — & — & — \\
TeCoNeRV$^\dagger$ & 41.3\,M & 65.2\,K & 0.047 &  $\text{\textbf{0.58}}\times\text{\textbf{10}}^3$
  & 25.15 & 0.721 & — & — & — & — \\
\rowcolor[gray]{0.9} \textbf{\methodname{}} ($N{=}1000$) & 47.7\,M & \textbf{64\,K} & \textbf{0.007} & $0.51\!\times\!10^3$ & \textbf{28.94} & \textbf{0.844} & \textbf{32.01} & \textbf{0.891} & \textbf{28.59} & \textbf{0.847} \\
\bottomrule
\end{tabular}
\caption{%
Comparison of our GS representation on Kinetics-400, SSV2, and UCF-101. Model \#Params denotes the feed-forward network size. Encoding time at per-GPU batch $B{=}8$. All models are trained on the K400 subset of 10,000 videos. $\dagger$ models trained \emph{and} evaluated at $640\times480$. Ours are trained at 256 $\times$ 256 and evaluated at higher resolutions zero-shot.}
\label{tab:sota}
\end{table*}

\begin{table}[t]
\centering
\setlength{\tabcolsep}{4pt}
\footnotesize
\begin{tabular}{l c c c}
\toprule
Method & \#Repr.\ Params & PSNR$\uparrow$ & SSIM$\uparrow$ \\
\midrule
\multicolumn{4}{l}{\textit{Feedforward methods at} 640 $\times$ 480 $\times$ 8}  \\
\midrule
TeCoNeRV& 4.9 \,M& 25.52 & 0.7103 \\
FastNeRV & 32.8\,M & 23.05 & 0.6491 \\
\rowcolor[gray]{0.9} \textbf{\methodname{}} ($N{=}1000$) & \textbf{4.8\,M} & \textbf{26.39} & \textbf{0.7157} \\
\bottomrule
\end{tabular}
\caption{%
Comparison with feedforward methods on UVG~\cite{mercat2020uvg} at $640\times480$, all frames (no temporal subsampling), zero-shot for all methods. \#Repr.\ Params is the total representation for a 600-frame UVG video. Our method stores $N{\times}8$ Gaussian parameters per frame ($\times600$), while other methods store the params for the INR (i.e. base params) per 8-frame clip ($\times75$ clips).  
}
\label{tab:uvg_hypernet}
\end{table}

\paragraph{\textbf{Comparison with Feed-Forward Methods.}}

Table~\ref{tab:sota} compares \methodname{} against feedforward INR methods at $256{\times}256$ resolution. At a representation \emph{smaller} than every prior feedforward method (64\,K vs.\ 85.6--139\,K), \methodname{} ($N{=}1000$) reaches 30.3\,dB on K400, 33.4\,dB on SSv2, and 29.8\,dB on UCF-101, surpassing the strongest prior method, FastNeRV~\cite{chen2024fast}, by \textbf{+2.9\,dB}, \textbf{+3.1\,dB}, and \textbf{+3.0\,dB} respectively. These gains are consistent across the in-distribution benchmark (K400) and zero-shot evaluation on SSv2 and UCF-101, and hold even though all methods use comparable model sizes (${\sim}$48\,M parameters). \methodname{} reaches this quality while encoding an 8-frame batch clip in ${\sim}7$\,ms (${\sim}1{,}193$\,FPS) and rasterizing at ${\sim}4{,}112$\,FPS ($0.51{\times}10^3$\,VPS). Decoding is slower than INR-based feedforward methods such as FastNeRV ($4.6{\times}10^3$\,VPS), which uses a specialized parallel decoder, since here the bottleneck is the differentiable Gaussian rasterizer rather than the network itself (see Appendix for details).

Furthermore, since $N$ is an explicit, tunable rate control for \methodname{}, this advantage is not limited to a single model, as we show in Sec.~\ref{sec:num-gaussians} (Table~\ref{tab:abl_points}), \methodname{} remains ahead of FastNeRV even at roughly a third of the representation size (27.89\,dB vs.\ 27.4\,dB on K400 at $N{=}500$, 32\,K per video), and scaling up to $N{=}3000$ (192\,K) raises quality further to 33.13\,dB / 36.23\,dB. 

Table~\ref{tab:sota} also evaluates rendering at $640{\times}480$. Because the predicted 2D Gaussians are \emph{continuous} primitives, the same predicted set is rasterized \emph{natively} at the target resolution with no re-encoding. Despite training only at $256{\times}256$ and being evaluated zero-shot, \methodname{} ($N{=}1000$) reaches 28.94\,dB on K400, outperforming FastNeRV (22.86\,dB) and TeCoNeRV (25.15\,dB)---both trained \emph{and} evaluated directly at $640{\times}480$---by wide margins. This indicates that the Gaussian representation transfers across resolutions better than implicit decoder outputs.

Table~\ref{tab:uvg_hypernet} extends this comparison to UVG at $640{\times}480$, evaluated zero-shot for all methods. \methodname{} ($N{=}1000$) reaches 26.39\,dB at 4.8\,M total parameters, surpassing FastNeRV (23.05\,dB) and TeCoNeRV (25.52\,dB) at a smaller representation.


\paragraph{\textbf{Comparison with Traditional Codec.}}

\begin{table}[t]
\centering
\setlength{\tabcolsep}{0.5mm}
\footnotesize
\begin{tabular}{l cc cc cc}
\toprule
 & \multicolumn{2}{c}{Speed (VPS)} & \multicolumn{2}{c}{K400} & \multicolumn{2}{c}{SSV2} \\
\cmidrule(lr){2-3} \cmidrule(lr){4-5} \cmidrule(lr){6-7}
Method & Enc$\uparrow$ & Dec$\uparrow$ & PSNR & SSIM & PSNR & SSIM \\
\midrule
\rowcolor[gray]{0.9} \textbf{\methodname{}} ($N{=}2000$) & \textbf{83} & \textbf{512} & 32.08 & 0.9054 & 35.15 & 0.9278 \\
H.265 (CRF30) & 62 & 385 & 32.34 & 0.9000 & 34.24 & 0.9170 \\
\midrule
\rowcolor[gray]{0.9} \textbf{\methodname{}} ($N{=}750$) & \textbf{179} & \textbf{537} & 29.25 & 0.8419 & 32.25 & 0.8790 \\
H.265 (CRF35) & 69 & 385 & 29.37 & 0.8413 & 31.40 & 0.8738 \\
\midrule
\rowcolor[gray]{0.9} \textbf{\methodname{}} ($N{=}500$) & \textbf{222} & \textbf{524} & 27.89 & 0.8064 & 30.83 & 0.8518 \\
H.265 (CRF40) & 81 & 385 & 26.50 & 0.7658 & 28.56 & 0.8183 \\
\bottomrule
\end{tabular}
\caption{%
Comparison with H.265 on K400 / SSV2 ($256^2$, 8-frame clips), grouped into matched-quality pairs. Within each pair we are faster at both encoding and decoding. Our feed-forward encoding is $1.3\text{--}2.7\times$ faster and rasterization decoding is ${\sim}1.3\times$ faster, with equal-or-higher fidelity. Speeds at $B{=}8$, ours on a single A100, H.265 on 8 CPUs.}
\label{tab:codec_compare}
\end{table}
 
Table~\ref{tab:codec_compare} compares \methodname{} against H.265, a widely-used block-based video codec, on $256^2$-resolution, 8-frame clips from K400 and SSV2. To enable a fair comparison across differing rate-distortion operating points, we group results into three matched-quality pairs, varying our Gaussian count $N \in \{2000, 750, 500\}$ against H.265 CRF settings. Across all three pairs, \methodname{} is faster than H.265 at \emph{both} encoding and decoding while achieving equal or higher reconstruction quality. Our feed-forward encoder is $1.3$--$2.7\times$ faster than H.265's encoder, with the gap widening as $N$ decreases (83 VPS at $N=2000$ up to 222 VPS at $N=500$, versus H.265's roughly flat 62--81 VPS). Decoding via rasterization is consistently ${\sim}1.3\times$ faster than H.265 decoding (512--537 vs.\ 385 VPS).

\section{Ablations}
\label{sec:ablations}

We ablate the key design choices in \methodname{}.

\subsection{Quality--Speed Trade-off with Gaussian Count.} 
\label{sec:num-gaussians}
Table~\ref{tab:abl_points} varies the number of Gaussians per frame from 250 to 3000. Reconstruction quality improves consistently with more Gaussians, gaining +7.31\,dB on K400 (25.82$\to$33.13\,dB) and +7.65\,dB on SSv2 (28.58$\to$36.23\,dB), with $N{=}3000$ yielding the best overall quality on both datasets. Since representation size scales linearly with $N$ (8 parameters per Gaussian), this trade-off directly trades storage for quality. Encoding cost scales similarly---per-batch encoding time grows from ${\sim}3.5$\,ms at $N{=}250$ to ${\sim}18.6$\,ms at $N{=}3000$---while rasterization speed remains roughly constant at $\sim$4000\,FPS (3972--4440) (see Appendix), and total training time is largely unaffected (19.36h--20.11h across the full range).

\begin{table*}[]
\centering
\footnotesize
\setlength{\tabcolsep}{0.5mm}
\begin{tabular}{cc c ccc cccc}
\toprule
 & & & \multicolumn{3}{c}{Speed} & \multicolumn{2}{c}{K400} & \multicolumn{2}{c}{SSV2} \\
\cmidrule(lr){4-6} \cmidrule(lr){7-8} \cmidrule(lr){9-10}
$N$ & Repr.\ \#Params  & Train (h) & Enc (s) & Enc (VPS/FPS) & Dec.\ (VPS/FPS) 
    & PSNR & SSIM & PSNR & SSIM \\
\midrule
250  & 2\,K  & 19.36 & \textbf{0.0035} & \textbf{286} / \textbf{2285} & \textbf{555 / 4440} & 25.82 & 0.7380 & 28.58 & 0.7999 \\
500  & 4\,K  & 19.51 & 0.0045 & 222 / 1779 & 524 / 4191 & 27.89 & 0.8064 & 30.83 & 0.8518 \\
750  & 6\,K  & 19.49 & 0.0056 & 179 / 1435 & 537 / 4293 & 29.25 & 0.8419 & 32.25 & 0.8790 \\
1000 & 8\,K  & 19.55 & 0.0067 & 149 / 1193 & 514 / 4112 & 30.29 & 0.8667 & 33.37 & 0.8983 \\
2000 & 16\,K & 20.06 & 0.0120 & \phantom{0}83 / 665\phantom{0} & 512 / 4093 & 32.08 & 0.9054 & 35.15 & 0.9278 \\
3000 & 24\,K & 20.11 & 0.0186 & \phantom{0}54 / 431\phantom{0} & 496 / 3972 & \textbf{33.13} & \textbf{0.9242} & \textbf{36.23} & \textbf{0.9430} \\
\bottomrule
\end{tabular}
\caption{Effect of the number of Gaussian primitives per frame $N$ on representation size, speed, and reconstruction quality, for 8-frame videos. Repr.\ \#Params is the per-frame representation size. Train (h) is wall-clock training time. Speed is measured on a single A100 at $B{=}8$, $T{=}8$, $256^2$ (K400-val): Enc is the per-batch network forward-pass time and Dec is the rasterization. VPS / FPS report throughput in videos/second and frames/second. PSNR and SSIM are reported on K400 and SSV2 validation videos.}
\label{tab:abl_points}
\end{table*}

\subsection{Adaptive Geometric Regularization.}
The geometric penalties ($\mathcal{L}_{erank}$ and $\mathcal{L}_{scale}$) address a known instability in Gaussian splatting optimization. Without any geometric constraints, the optimization process learns to generate highly anisotropic, ``needle-like'' Gaussians to fit sharp edges. While this initially minimizes the reconstruction loss ($\mathcal{L}_{recon}$), the model eventually over-accumulates these needles. This triggers a sudden, sharp collapse in performance, often occurring unexpectedly mid-way through the training. We find that this instability is highly dependent on task complexity and model capacity and requires excessive hyper parameter searching for different gaussian points and configurations to stabilize and increase the performance. We address this with an EMA-based adaptive approach to calibrate the regularization uniquely for each model. 

\begin{figure*}[t]
    \centering
    \footnotesize
    \begin{subfigure}{0.49\textwidth}
        \centering
        \includegraphics[width=\linewidth]{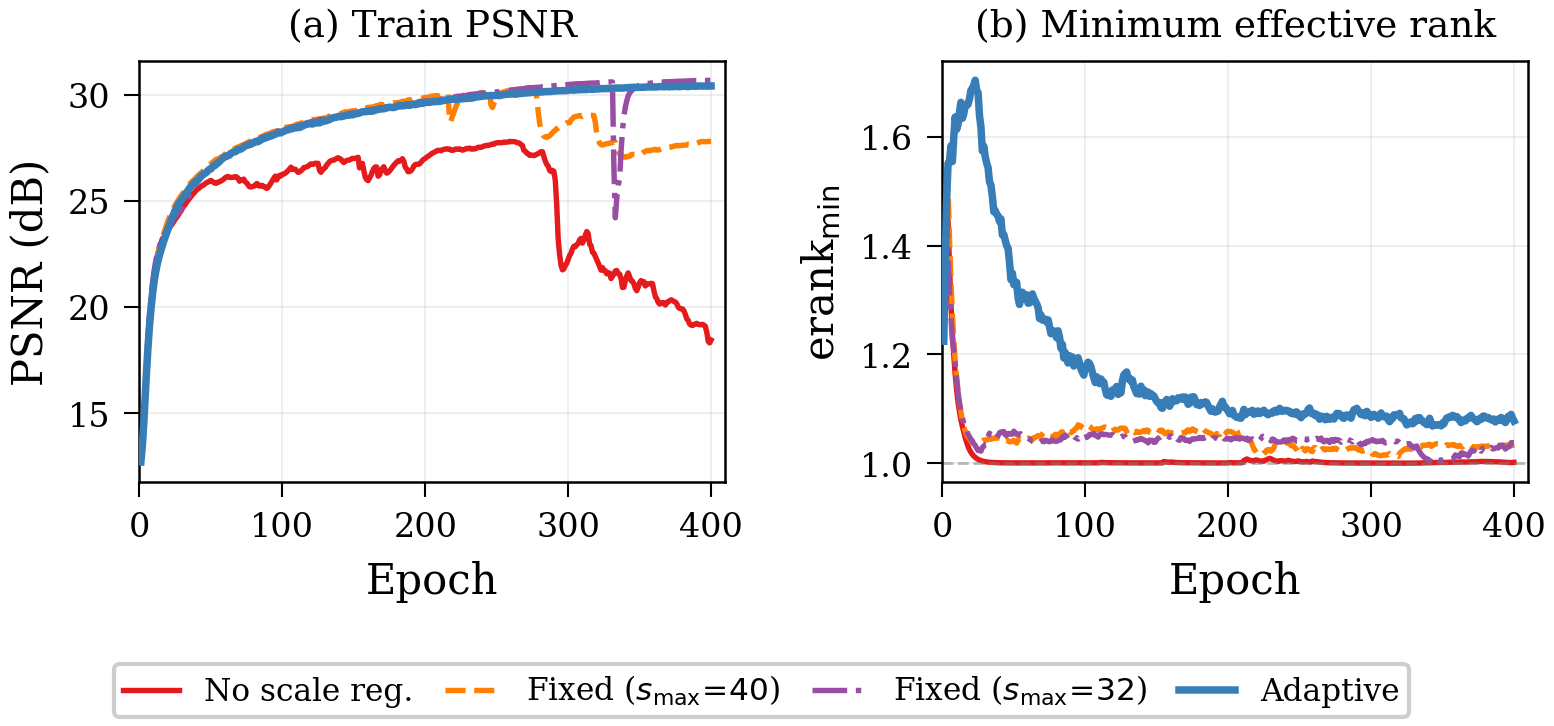}
        \caption{1000 Gaussian Points}
        \label{fig:reg_stability}
    \end{subfigure}
    \begin{subfigure}{0.49\textwidth}
        \centering
        \includegraphics[width=\linewidth]{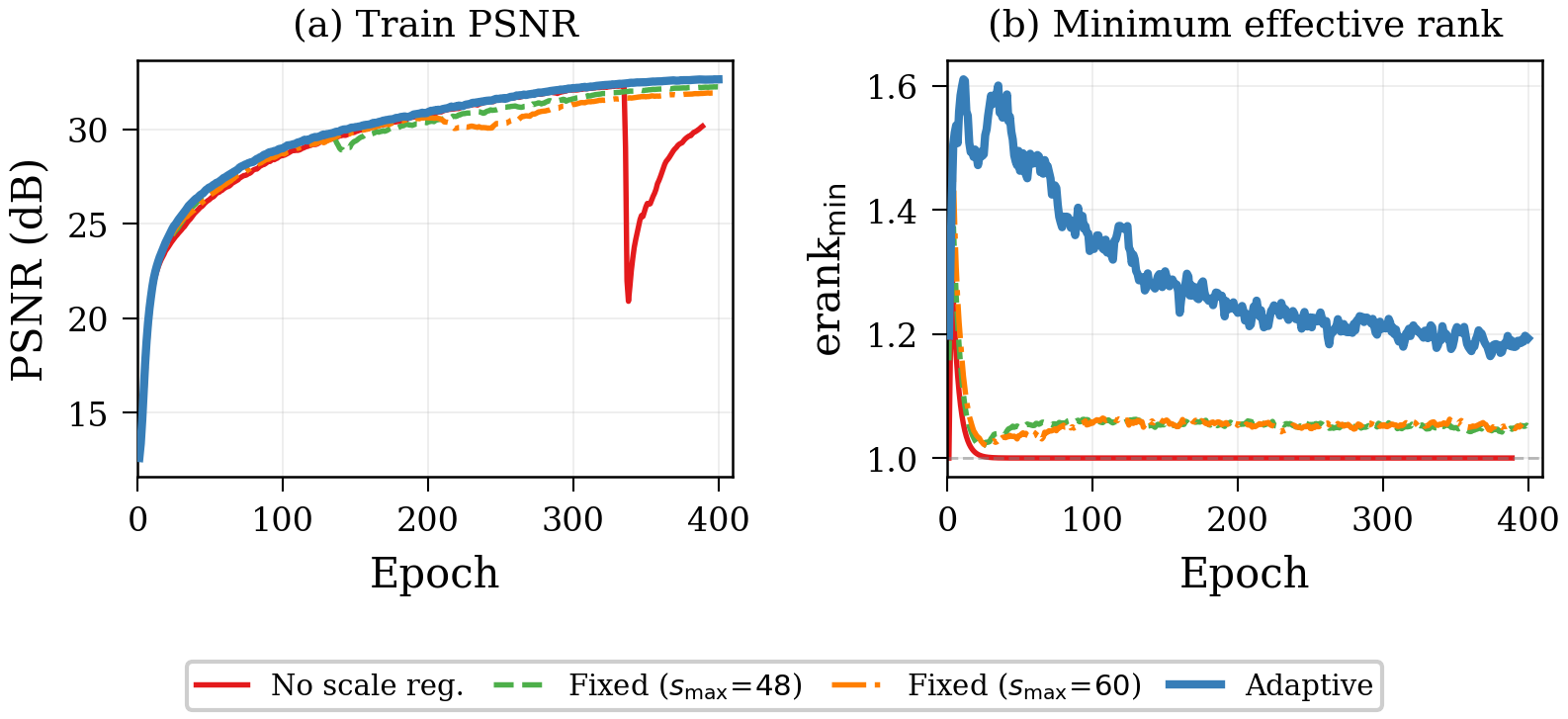}
        \caption{2000 Gaussian points}
        \label{fig:reg_rank}
    \end{subfigure}

    \caption{Training Stability. We analyze the training PSNR and minimum effective rank over 400 epochs across different regularization configurations.
    Without regularization, PSNR collapses as Gaussians degenerate into ``needles'' ($erank \to 1$). While fixed thresholds (ex. $\tau_{erank}=1.1$ and different $s_{max}$) can achieve high peak performance, they are sensitive to model capacity and require extensive hyperparameter searching.
    }
    \label{fig:adaptive_reg}
\end{figure*}

Figure~\ref{fig:adaptive_reg} compares fixed-threshold regularization, in which the quality is sensitive to the choice of $\tau_{\text{erank}}$ and $s_{\max}$, with our adaptive EMA-based approach (Section~\ref{sec:method}). Stricter settings (higher $\tau_{\text{erank}}$ and lower $s_{\max}$) are able to recover from collapses, but over-constrain useful elliptical Gaussians, while more relaxed settings risks more instability. The adaptive regularization, initialized with the stricter setting, gradually constrains thresholds when outliers emerge and relaxes them otherwise, consistently stabilizing training. 

\subsection{Q-Transformer Type and Model Capacity.}
Table~\ref{tab:abl_arch} isolates the role of the Q-Transformer and of overall capacity. We observe that replacing the self-attention Q-Transformer with an MLP that directly regresses Gaussian parameters collapses quality to 22.98\,dB---9.1\,dB below a self-attention Q-Transformer at the same $N{=}2000$ (30.29\,dB)---with LPIPS and tLP degrading as well. This shows that query-based aggregation, in which Gaussian tokens attend over the contextualized frame features, is what makes Gaussian prediction work effectively, rather than raw network capacity. We also observe that reducing the hidden width ($d{=}512$) costs only $-0.78$\,dB at roughly half the parameters, and trimming Q-Transformer depth (ST2/Q3, ST2/Q2) trades quality smoothly for a smaller, faster model. 

\begin{table}[h]
{
\setlength{\tabcolsep}{1mm}
\footnotesize
\begin{tabular}{@{}l cc c cc c@{}}
\toprule
 & HyperN.\ \#P  & Enc (s) & PSNR$\uparrow$ & SSIM$\uparrow$ & LPIPS$\downarrow$ & tLP$\downarrow$ \\
\midrule
\multicolumn{7}{@{}l}{\textit{Full model (reference): ST2/Q4, $d{=}768$, self-attn, $N{=}2000$}} \\
\midrule
ST2/Q4            & 48.4\,M & 0.0120 & \textbf{32.08} & \textbf{0.9054} & \textbf{0.1584} & \textbf{0.1558} \\
\midrule
\multicolumn{7}{@{}l}{\textit{Network depth} (vary ST/Q layers; $d{=}768$, $N{=}2000$)} \\
\midrule
ST1/Q2       & 26.1\,M & 0.0064 & 29.62 & 0.8560 & 0.2282 & 0.2192 \\
ST2/Q2       & 35.1\,M & 0.0070 & 31.04 & 0.8855 & 0.1870 & 0.1851 \\
ST2/Q3       & 41.8\,M & 0.0097 & 31.85 & 0.9017 & 0.1621 & 0.1599 \\
\midrule
\multicolumn{7}{@{}l}{\textit{Hidden width} (vary $d$; ST2/Q4, $N{=}2000$)} \\
\midrule
$d{=}512$       & 23.5\,M & 0.0072 & 31.30 & 0.8911 & 0.1816 & 0.1790 \\
\midrule
\multicolumn{7}{@{}l}{\textit{Decoder type} (MLP vs.\ self-attn; $d{=}768$, $N{=}2000$)} \\
\midrule
MLP             & 39.6\,M & 0.0060 & 22.98 & 0.6459 & 0.6226 & 0.6287 \\
\bottomrule
\end{tabular}}
\caption{
Architecture ablation on K400. We vary network depth, hidden width, and Q-Transformer type relative to our full model (ST2/Q4, $d{=}768$, self-attention). 
}
\label{tab:abl_arch}
\end{table}

\section{Conclusion}
\label{sec:conclusion}
In this work, we introduced \methodname{}, a fast and generalizable network for representing videos with Gaussian Splatting. By predicting explicit geometric primitives directly from pixels, \methodname{} leverages an inductive bias that decomposes video reconstruction into interpretable subproblems: position, scale, rotation, and color are regressed independently for a fixed set of Gaussians per frame. This design enables real-time encoding while retaining the high visual fidelity of Gaussian Splatting representations.

\bibliography{aaai2027}


\appendix

\clearpage

\section{Quantization Rate--Distortion Curves}
\begin{figure}[th]
\centering
\begin{subfigure}{0.39\textwidth}
  \includegraphics[width=\linewidth]{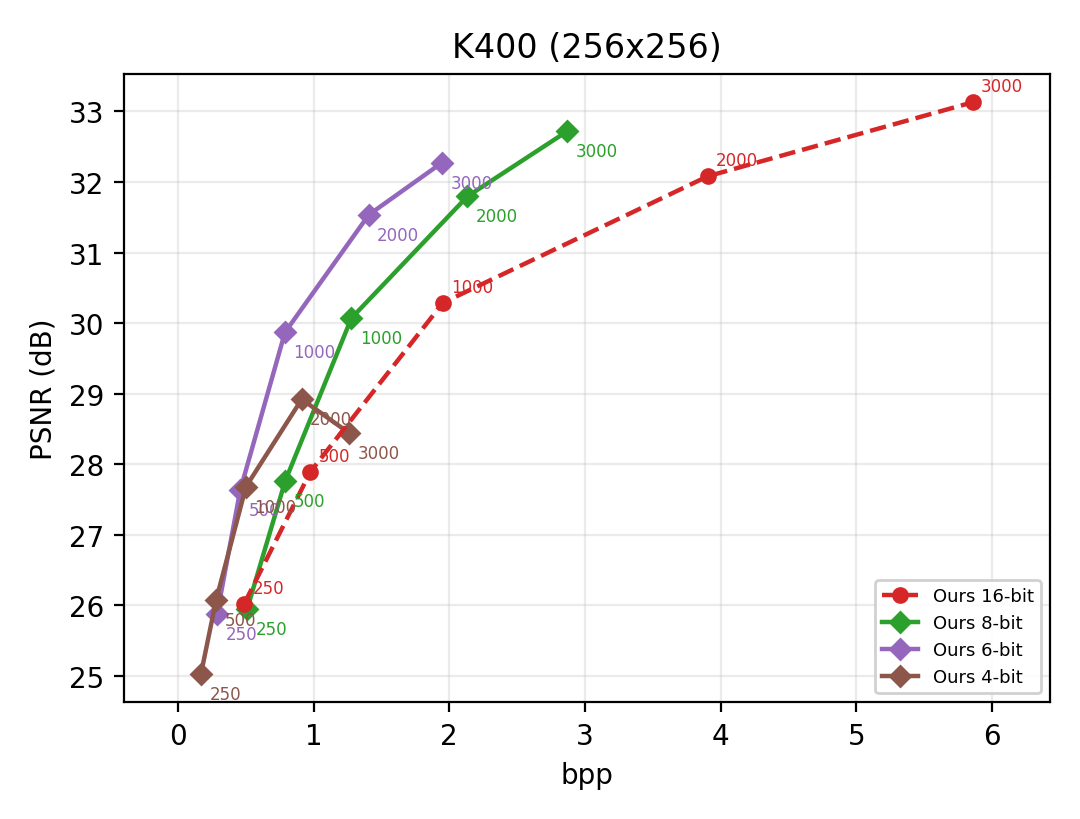}
\end{subfigure}
\begin{subfigure}{0.39\textwidth}
  \includegraphics[width=\linewidth]{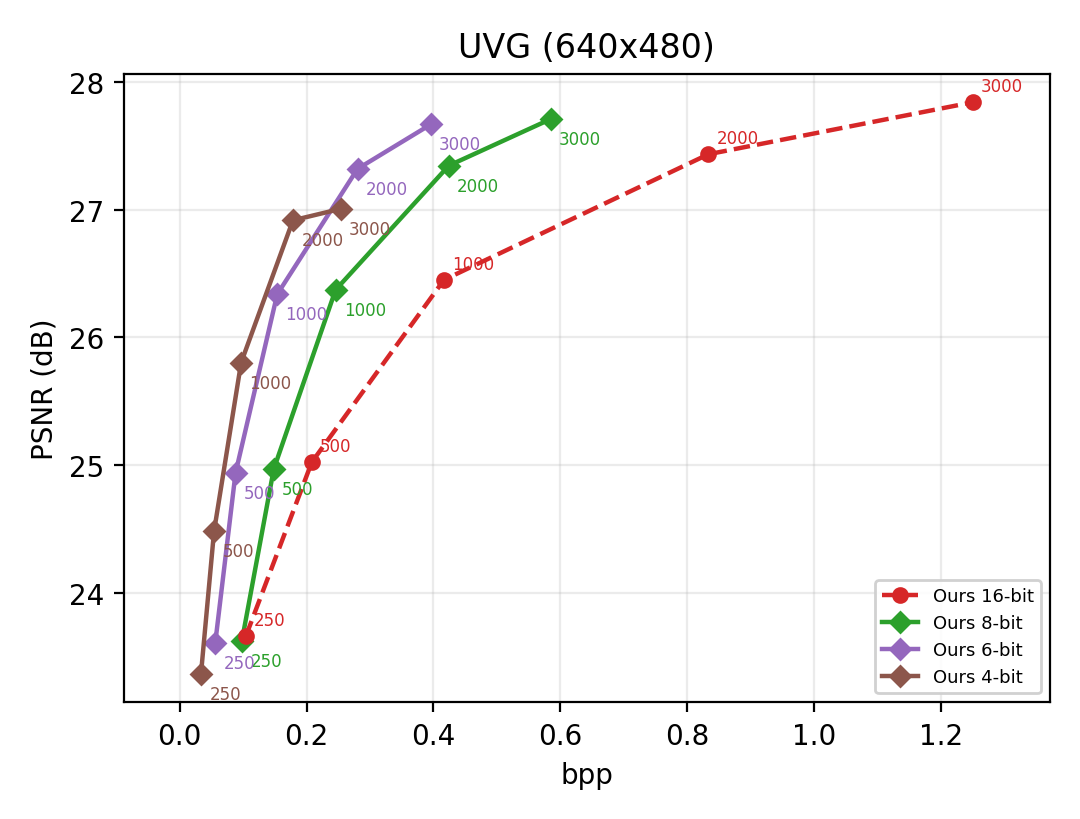}
\end{subfigure}

\caption{Rate--Distortion Curves. We evaluate post-training compression while varying the model capacity ($N{=}250\text{--}3000$). We apply quantization plus entropy (ANS) coding to the predicted Gaussians with cross-frame residual quantization---an intra-coded anchor frame plus quantized P-frame deltas, swept at 8/6/4-bit precision.}
\label{fig:rd_quant}
\end{figure}

Since each Gaussian stores 8 FP16 scalars, the per-frame bitrate is $\text{bpp} = (N \times 8 \times 16)/(H \times W)$, so $N$ is a directly interpretable control for rate--distortion. Figure~\ref{fig:rd_quant} plots this trade-off on K400 and UVG, showing that \methodname{} traces a favorable rate--distortion curve: PSNR grows sub-linearly with bitrate, following the expected diminishing-returns behavior as $N$ increases.

To further compress the representation, we quantize the Gaussian parameters and entropy-code them with an asymmetric numeral system (ANS), plotting PSNR against the resulting bits-per-pixel. Our codec uses a cross-frame residual approach that exploits the temporal consistency of our prediction: since the decoder's weight tokens place the $i$-th Gaussian consistently across frames, we treat frame~0 as an intra-coded anchor and entropy-code only the quantized residuals of subsequent frames relative to it. We report results at anchor/residual precisions of 8, 6, and 4 bits, showing that our quantized models improve along the RD curve relative to our unquantized FP16 baseline.

\section{Qualitative Results}
\label{sec:qualitative}

\begin{figure*}[h]
  \centering
  \includegraphics[width=0.95\linewidth]{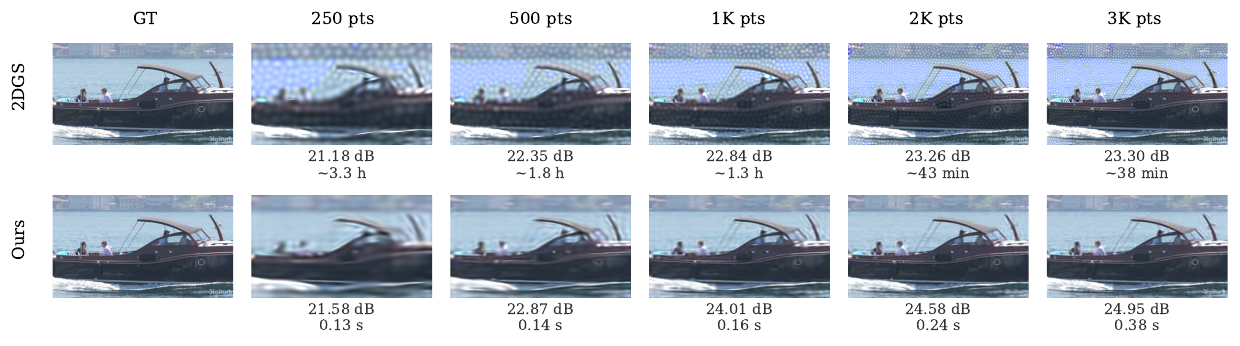}
  \includegraphics[width=0.95\linewidth]{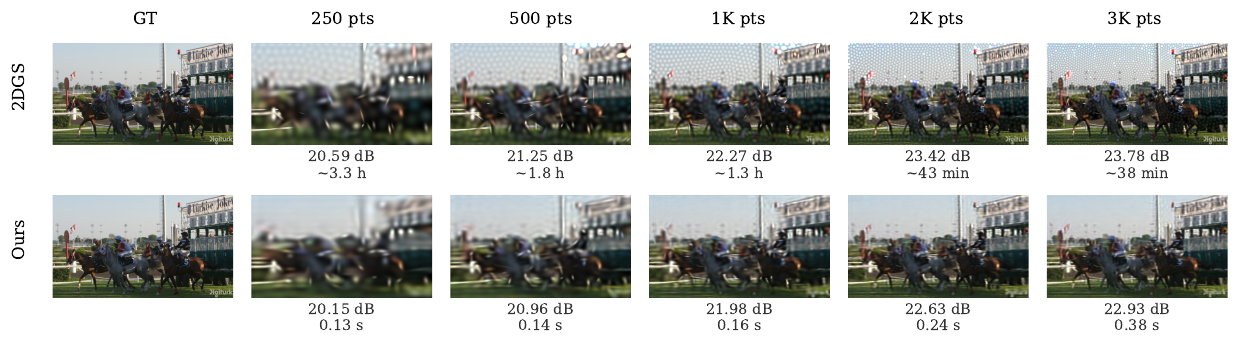}
  \includegraphics[width=0.95\linewidth]{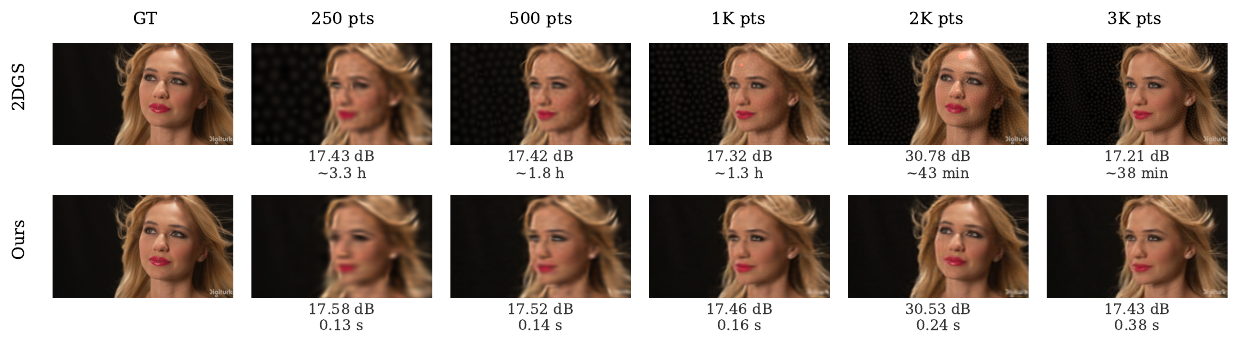}

  \caption{\textbf{Qualitative comparison on the UVG sequences at different Gaussian points.} We compare per-frame reconstructions from 2DGS~\cite{zhang2024gaussianimage}, which optimizes Gaussian parameters per frame at (1280$\times$720) and our network, which predicts them in a single forward pass, zero-shot at higher resolution. PSNR (dB) is reported for each frame alongside the average per-video encoding time for UVG videos.}
  \label{fig:qualitative_uvg}
\end{figure*}

Figure~\ref{fig:qualitative_uvg} compares \methodname{} against 2DGS~\cite{zhang2024gaussianimage} on UVG at $1280{\times}720$ across Gaussian counts from 250 to 3K. Both methods use the same 2D Gaussian rasterization pipeline, but they differ in how the Gaussian parameters are obtained: 2DGS optimizes them per frame for up to 50{,}000 iterations, whereas \methodname{} predicts them in a single forward pass. To make the comparison fair, we compare both methods at matched reconstruction quality. For each Gaussian count we take the 2DGS encoding whose average PSNR matches \methodname{} on the encoding-time--quality curves in Fig.~\ref{fig:davis-uvg-encoding-psnr}---equivalently, the point where 2DGS reaches \methodname{}'s PSNR at higher encoding time. Under sequential per-frame optimization, this corresponds to total per-video costs ranging from ~38 min to ~3.3 h for 2DGS, compared with $<$1 s for ours. This matched-PSNR protocol isolates differences in reconstruction \emph{appearance} at equal quality.

At matched PSNR, the two methods still degrade differently at lower Gaussian counts. 2DGS tends to produce structured artifacts---visible as sharp discontinuities and tiling patterns---whereas \methodname{} degrades with more uniform blur. This reflects how the network learns a smoother prediction averaged across its training distribution, while per-frame optimization can overfit local structure at the cost of spatial coherence. As the Gaussian count increases, both methods improve and their outputs become more similar, though their characteristic failure modes remain visible even at comparable PSNR.

\begin{figure*}[h]
  \centering
  \includegraphics[width=\textwidth]{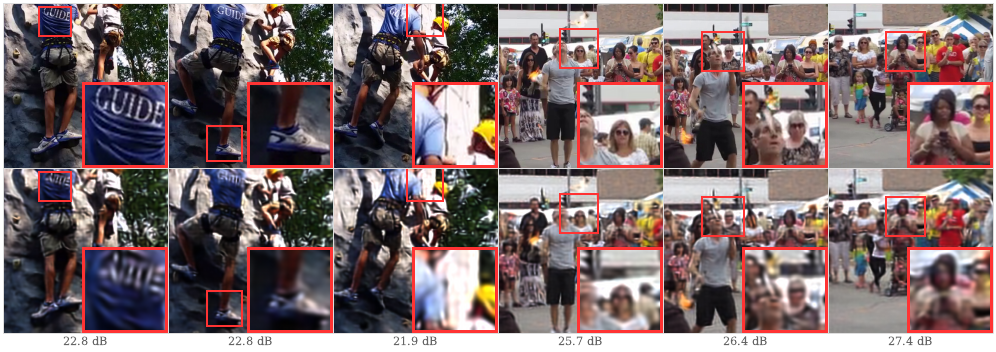}
  \caption{\textbf{Qualitative reconstruction results on Kinetics-400.} The top row shows ground truth frames and the second row shows our reconstructions across three frames. Per-frame PSNR (dB) is reported below each video.}
  \label{fig:qualitative_k400}
\end{figure*}

Figure~\ref{fig:qualitative_k400} shows reconstruction examples from \methodname{} on Kinetics-400 across three temporal frames per video. The model recovers scene structure, color, and some texture showing consistent PSNR across frames within each clip.

\section{Additional Ablations}

\subsection{ST-Transformer--Q-Transformer Depth Allocation.}
Table~\ref{tab:abl_depth} varies how 6 total transformer layers are split between the ST-Transformer and the Q-Transformer. A Q-Transformer-only configuration (0 ST / 6 Q layers) achieves 28.35\,dB. Introducing even a single ST-Transformer layer (1/5 split) improves quality by +1.3\,dB, and performance continues to increase up to a 3/3 split (30.39\,dB). Thus, separating spatio-temporal contextualization from query-based regression helps build richer spatiotemporal features that the Q-Transformer can more effectively translate into Gaussian parameters. Our default configuration uses a 2/4 split (30.29\,dB), which achieves a good balance between quality and parameter efficiency.

\begin{table}[t]
\centering

\setlength{\tabcolsep}{6pt}
\footnotesize
\begin{tabular}{cc c c c}
\toprule
ST depth & Q depth & HyperN.\ \#P & PSNR$\uparrow$ & SSIM$\uparrow$ \\
\midrule
0 & 6 & 42.9M & 28.35 & 0.7923 \\
1 & 5 & 45.3M & 29.67 & 0.8511 \\
2 & 4 & 47.7M & 30.29  & 0.8395\\
3 & 3 & 50.0M & \textbf{30.39} & \textbf{0.8656} \\
\bottomrule
\end{tabular}
\caption{ST-Transformer--Q-Transformer depth allocation on K400. A total of 6 transformer layers is distributed between the ST-Transformer and the Q-Transformer. All models use 1000 Gaussians, 8 frames, and factorized spatiotemporal attention.}
\label{tab:abl_depth}
\end{table}

\subsection{Q-Transformer Attention Type}
\label{appendix:decoder_type}
We compare three Q-Transformer variants, all using ST2/Q4 with the factorized spatio-temporal architecture and $N=2000$ Gaussians. In the default \emph{self-attention} variant, per-frame patch tokens from the ST-Transformer are concatenated with the learnable Gaussian tokens and processed jointly through self-attention layers. Each Gaussian token can attend to all patch tokens within its frame as well as to other Gaussian tokens. The \emph{cross-attention} variant instead separates the two token sets such that the Gaussian tokens serve as queries that cross-attend to the full ST-Transformer output across \emph{all} frames (not just the current frame), followed by a feedforward block. The \emph{cross-attention + self-attention} variant adds a self-attention layer among the Gaussian tokens after each cross-attention layer, allowing Gaussian tokens to also exchange information with each other. We observe that the self-attention variant outperforms cross-attention by +0.54\,dB.
\begin{table}[h]
\centering
\footnotesize
\begin{tabular}{l c c}
\toprule
Decoder type & PSNR$\uparrow$ & SSIM$\uparrow$ \\
\midrule
Cross-attention              & 31.54 & 0.8945 \\
Cross-attention + self-attention & 31.95  &  0.9023 \\
Self-attention       & \textbf{32.08} & \textbf{0.9054} \\
\bottomrule
\end{tabular}
\caption{Comparison of Q-Transformer attention mechanisms on K400 
at $256\times256$ with $N=2000$, ST2/Q4.}
\label{tab:decoder_type}
\end{table}

\subsection{Temporal Attention Window Size}
\label{appendix:temporal_window}
Table~\ref{tab:temporal_window} ablates the temporal window size~$w$ in the factorized spatio-temporal architecture.
The window size controls the scope of temporal attention. Each ST-Transformer layer performs spatial attention within individual frames, followed by temporal attention restricted to non-overlapping windows of $w$ consecutive frames.
Setting $w=8$ (the default) enables full temporal attention across all input frames, while $w=2$ limits each token to attending only within pairs of adjacent frames.
Performance improves incrementally with window size, from 30.04\,dB at $w=2$ to 30.29\,dB at $w=8$, indicating that long-range temporal context benefits per-frame Gaussian prediction, albeit minimally.

\begin{table}[th]
\centering
\footnotesize
\begin{tabular}{c c c c c c}
\toprule
Window $w$ & PSNR$\uparrow$ & SSIM$\uparrow$ & LPIPS$\downarrow$ & $\Delta$-PSNR$\uparrow$ & tLP$\downarrow$ \\
\midrule
2 & 30.04 & 0.8610 & 0.216 & 32.30 & 0.213 \\
4 & 30.26 & 0.8656 & 0.211 & 32.49 & 0.207 \\
8 & \textbf{30.29} & \textbf{0.8667} & \textbf{0.208} & \textbf{32.51} & \textbf{0.204} \\
\bottomrule
\end{tabular}
\caption{Effect of temporal attention window size on K400 at
$256\times256$ with $N=1000$, 8 input frames.}
\label{tab:temporal_window}
\end{table}

\subsection{Number of Frames.}
Table~\ref{tab:ablation_frames_full} evaluates video at 4, 8, and 16 uniformly sampled frames for 1000 Gaussians. We observe that the model maintains highly stable performance as the sequence length increases. Since each frame is rendered from its own independent set of Gaussians, the temporal length primarily challenges the encoder's ability to contextualize extended sequences without degrading. While 8 frames yield a slight peak in PSNR, extending the sequence to 16 frames achieves the best results across all perceptual (SSIM, LPIPS) and temporal fidelity ($\Delta$-PSNR, tLP) metrics.

\begin{table*}[h]
\centering
\footnotesize
\setlength{\tabcolsep}{4pt}
\begin{tabular}{@{}cc|ccccc|ccccc@{}}
\toprule
 & & \multicolumn{5}{c|}{K-400} & \multicolumn{5}{c}{SSv2} \\
\cmidrule(lr){3-7} \cmidrule(lr){8-12}
$N$ & $T$ &
PSNR$\uparrow$ & SSIM$\uparrow$ & LPIPS$\downarrow$ &
$\Delta$-PSNR$\uparrow$ & tLP$\downarrow$ &
PSNR$\uparrow$ & SSIM$\uparrow$ & LPIPS$\downarrow$ &
$\Delta$-PSNR$\uparrow$ & tLP$\downarrow$ \\
\midrule
\multirow{3}{*}{1000}
 & 4  & 30.23 & 0.8653 & 0.2097 & 32.2 & 0.2094 &
        33.29 & 0.8963 & 0.1735 & 34.8 & 0.1646 \\
 & 8  & \textbf{30.29} & 0.8667 & 0.2085 & 32.5 & 0.2040 &
        \textbf{33.37} & 0.8983 & 0.1708 & 35.2 & 0.1583 \\
 & 16 & 30.24 & \textbf{0.8669} & \textbf{0.2071} & \textbf{32.7} & \textbf{0.2014} &
        33.32 & \textbf{0.8988} & \textbf{0.1694} & \textbf{35.8} & \textbf{0.1555} \\
\bottomrule
\end{tabular}%
\caption{\textbf{Number of Frames}.
$\Delta$-PSNR denotes temporal fidelity and tLP denotes temporal LPIPS.}
\label{tab:ablation_frames_full}
\end{table*}

\subsection{Comparison with Per-Video Optimization (MS-SSIM).}
\begin{figure}[h]
  \centering
  \begin{subfigure}[b]{\linewidth}
    \centering
    \includegraphics[width=0.9\linewidth]{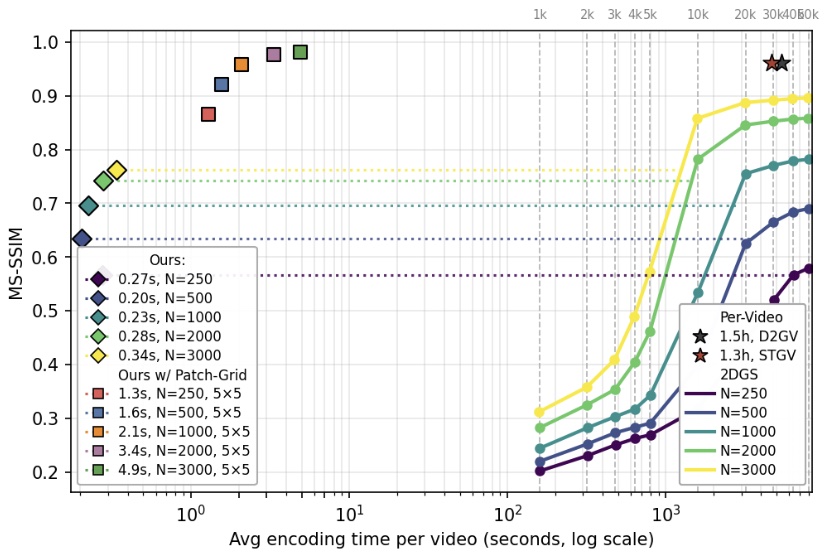}
    \caption{Davis 2016 MS-SSIM at 720p, $\sim$82 frames/video.}
    \label{fig:davis-encoding-psnr}
  \end{subfigure}
  \begin{subfigure}[b]{\linewidth}
    \centering
    \includegraphics[width=0.9\linewidth]{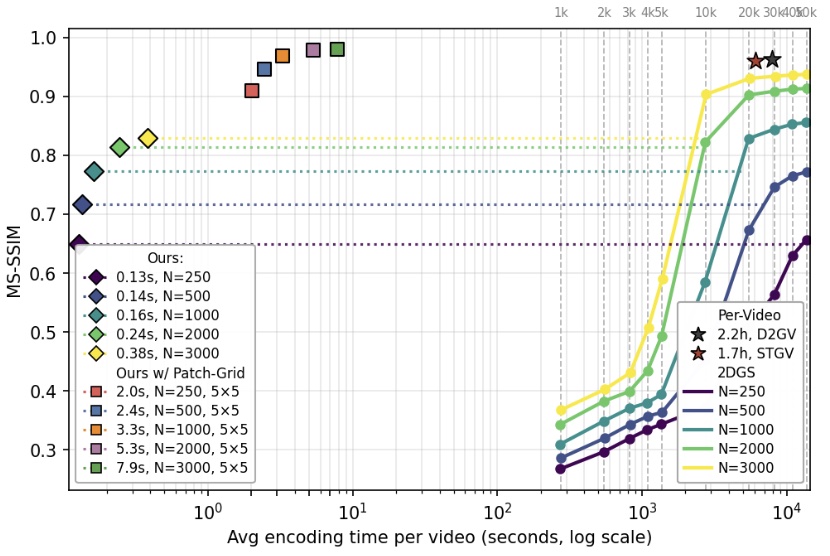}
    \caption{UVG MS-SSIM at 720p. $\sim$139 frames/video.}
    \label{fig:second-figure-label}
  \end{subfigure}
  \caption{ Each curve shows 2DGS~\cite{zhang2024gaussianimage} trained \emph{per-frame} with a fixed number of Gaussians (250--3000). The curve plots reconstruction quality, measured in MS-SSIM, against the average \emph{per-video} encoding time at different training iterations ($1k-50k$). 
  } 
  \label{fig:davis-uvg-encoding-msssim}
\end{figure}

Figure~\ref{fig:davis-uvg-encoding-msssim} reports the same comparison as Figure~\ref{fig:davis-uvg-encoding-psnr}, using MS-SSIM in place of PSNR to evaluate \methodname{} against its non-amortized counterparts in terms of encoding speed and reconstruction quality, evaluated zero-shot at 720p on DAVIS 2016 and UVG. As before, each 2DGS curve corresponds to a fixed Gaussian count $N$ optimized independently per frame for a given number of iterations (up to 50k), with more iterations yielding a tighter per-frame fit and higher MS-SSIM.
We also evaluate with patch-grid variants encoding $g_c\times g_r$ views per frame. Encoding time for our methods is the GPU pipeline consisting of a single forward pass plus Gaussian rasterization, batched at $B{=}8$. The encoding time for the compared methods is the training time.

\section{Inference at Higher Resolution}
\begin{figure}[t]
\centering
  \includegraphics[width=\linewidth]{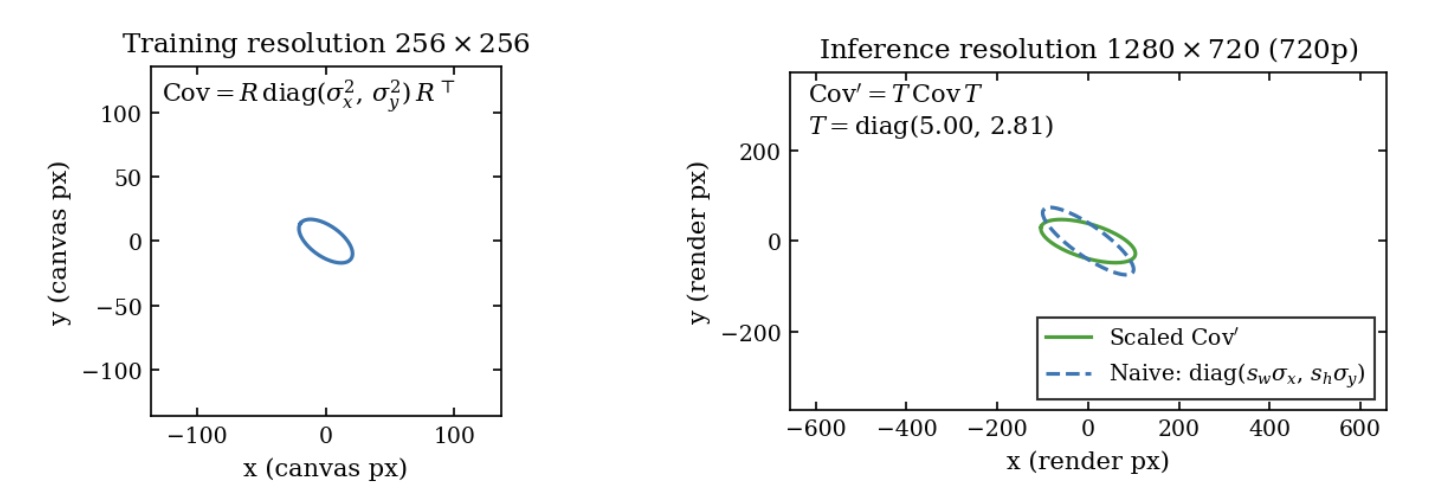}
\caption{Rendering at higher resolution stretches predicted Gaussian covariances via
  $\mathrm{Cov}'=T\,\mathrm{Cov}\,T$. Left: Decoder predicts scales $\sigma_x$, $\sigma_y$, and rotation $\theta$ per Gaussian on the $256^2$ canvas. Right: Render at $1280\times720$ stretches $\times5.00$ in width and $\times2.81$ in height. Exact vs.\ naive covariance under anisotropic stretch when rasterizing at 720p.}
  \label{fig:anistropic_scaling}
\end{figure}
The Gaussians predicted on a $H_c \times W_c$ canvas are rasterized directly at target resolution $H \times W$ by stretching image coordinates with $T = \mathrm{diag}(s_w, s_h)$, where $s_w = W/W_c$ and $s_h = H/H_c$. Each Gaussian is defined on the canvas by scales $(\sigma_x, \sigma_y)$ and rotation $\theta$, giving 
\begin{equation}
\mathrm{Cov} = R \mathrm{diag}(\sigma_x^2, \sigma_y^2) R^\top
\end{equation}
with $R = \left(\begin{array}{cc}\cos\theta & \sin\theta \\ -\sin\theta & \cos\theta\end{array}\right)$. Under anisotropic upscaling $(s_w \neq s_h)$, the projected covariance is 
\begin{equation}
\mathrm{Cov}' = T \mathrm{Cov} T
\end{equation}
which cannot be obtained by simply scaling $\sigma_x, \sigma_y$ and keeping $\theta$ unchanged. We therefore compute the Cholesky factor $\mathrm{Cov}' = LL^\top$ with $L = \left(\begin{array}{cc}L_{00} & 0 \\ L_{10} & L_{11}\end{array}\right)$ and pass $[L_{00}, L_{10}, L_{11}]$ to the 2D Gaussian projector. Writing $c=\cos\theta$, $s=\sin\theta$, and $\sigma_x^2, \sigma_y^2$ for the canvas scales, the entries of $\mathrm{Cov}'$ are 
\begin{eqnarray} 
a &=& s_w^2(c^2\sigma_x^2 + s^2\sigma_y^2), \\ 
b &=& s_w s_h s c(\sigma_y^2 - \sigma_x^2), \\ 
c' &=& s_h^2(s^2\sigma_x^2 + c^2\sigma_y^2), 
\end{eqnarray} 
from which the Cholesky elements follow as $L_{00}=\sqrt{a}$, $L_{10}=b/L_{00}$, and $L_{11}=\sqrt{c' - L_{10}^2}$. When $s_w = s_h$, this reduces to uniform scaling and the standard scale--rotation path is used instead.

\section{Implementation Details}
\label{appendix:implementation}

\subsection{Decoding Speeds and Rasterization Bottlenecks}
\label{appendix:decoding_speed}
\methodname{} achieves fast encoding and decoding speeds, driven by targeted batching of the 2DGS rasterization pipeline \cite{zhang2024gaussianimage}. The 2DGS rendering pipeline is split into two primary stages: geometric projection and tile-based accumulation. To optimize the decoding speed, we batch the first stage---the geometric transformations that map predicted scales, rotations, and 2D means into inverse covariance matrices ($\boldsymbol{\Sigma}_i^{-1}$) and frame-space coordinates are computed fully batched. This batching is done by flattening the batch and point dimensions, all $B \times N$ Gaussians are projected across all frames in the batch, thereby maximizing parallel throughput. The tile-based accumulation stage loops over the batch dimension at the frame level, trading some of the parallelism used in the fully batched projection stage — leaving room for further throughput gains relative to a fully parallelized decoder.

\subsection{Adaptive Regularization Details}
\label{appendix:adaptive_reg}

Table~\ref{tab:adaptive_reg_hparams} lists the hyperparameters of our adaptive regularization approach. The \emph{EMA momentum}~$m$ controls how quickly the thresholds respond to distributional shifts, providing stable estimates while still adapting within a few epochs.
The \emph{initial thresholds} ($\tau_{\text{erank}}^{(0)}$, $s_{\max}^{(0)}$) set conservative starting points that prevent early-training collapse before the EMA has converged.
The \emph{floor} $\tau_{\text{floor}}$ prevents the erank threshold from drifting toward~1 under collective degeneration.
Similarly, the \emph{ceiling} $s_{\text{ceil}}$ prevents the scale threshold from growing unbounded if the distribution's upper percentile drifts upward, which can trigger sudden collapse as Gaussians start spanning the entire image.
For $N=3000$, we relax the regularization: loss weights are halved, the erank floor and initial threshold are set near unity, and more extreme percentiles are used.
With a larger Gaussian budget, each primitive covers a smaller image region, so tighter shape constraints are unnecessary and overly restrictive regularization can hurt reconstruction quality.

\begin{table}[th!]
\centering
\footnotesize
\begin{tabular}{lccc}
\toprule
Parameter & Symbol & $N \leq 2000$ & $N = 3000$ \\
\midrule
Erank loss weight          & $\lambda_{\text{erank}}$        & 0.01   & 0.005  \\
Scale loss weight          & $\lambda_{\text{scale}}$        & 0.01   & 0.005  \\
Initial erank threshold    & $\tau_{\text{erank}}^{(0)}$     & 1.05   & 1.001  \\
Initial scale threshold    & $s_{\max}^{(0)}$                & 32.0   & 32.0   \\
EMA momentum               & $m$                             & 0.99   & 0.98   \\
Erank floor                & $\tau_{\text{floor}}$           & 1.005  & 1.001  \\
Scale ceiling              & $s_{\text{ceil}}$               & 80.0   & 80.0   \\
Erank percentile           & $p_{\text{erank}}$              & 0.01   & 0.0015 \\
Scale percentile           & $p_{\text{scale}}$              & 0.99   & 0.999  \\
\bottomrule
\end{tabular}
\caption{Adaptive geometric regularization hyperparameters.}
\label{tab:adaptive_reg_hparams}

\end{table}

\subsection{Training Configuration}
Architecture, optimization, and data hyperparameters—including learning rate schedule, and augmentation strategy—are reported in Table~\ref{tab:shared_hparams}.

\begin{table}[!th]
\centering
\footnotesize
\begin{tabular}{lc}
\toprule
Hyperparameter & Value \\
\midrule
\multicolumn{2}{l}{\textit{Architecture}} \\
Transformer dimension $d$              & 768 \\
Attention heads / head dim             & 12 / 64 \\
Feedforward hidden dim                 & 2800 \\
Total transformer depth                & 6 \\
Patch size                             & $16\times16$ \\
Tubelet size                           & 1 \\
Input resolution                       & $256\times256$ \\
ST-Transformer Layers & 2 \\
Q-Transformer Layers &  4 \\
\midrule
\multicolumn{2}{l}{\textit{Training}} \\
Optimizer                              & LAMB \\
Learning rate                          & $10^{-3}$ \\
LR schedule                            & Cosine \\
Minimum LR ratio                       & 0.1 \\
Warmup epochs                          & 0 \\
Epochs                                 & 400 \\
Batch size                             & 32 \\
Mixed precision                        & FP16 \\
Gradient clip norm                     & 1.0 \\
\midrule
\multicolumn{2}{l}{\textit{Data}} \\
Training set                           & Kinetics-400 (25\%/class) \\
Train augmentation                     & RandomCrop + HFlip \\
Test augmentation                      & None (CenterCrop) \\
\bottomrule
\end{tabular}
\caption{Training hyperparameters shared across all main experiments.}
\label{tab:shared_hparams}

\end{table}

\end{document}